\documentclass[sigconf,screen]{acmart}

\usepackage{amsmath}
\usepackage{booktabs}
\usepackage{graphicx}
\usepackage{xcolor}
\usepackage{enumitem}
\usepackage{placeins}

\setcopyright{cc}
\setcctype{by}
\acmDOI{10.1145/3843282.3844438}
\acmYear{2026}
\copyrightyear{2026}
\acmISBN{979-8-4007-2985-0/2026/10}
\acmConference[AgenticDev '26]{Proceedings of the 1st International Workshop on Agentic AI for Next-Generation Software Development}{October 12--16, 2026}{Munich, Germany}
\acmBooktitle{Proceedings of the 1st International Workshop on Agentic AI for Next-Generation Software Development (AgenticDev '26), October 12--16, 2026, Munich, Germany}
\acmSubmissionID{asews26agenticdevmain-p114-p}
\received{2026-07-23}
\received[accepted]{2026-08-20}
\graphicspath{{figures/}{../experiments/figures/v07/}{../experiments/figures/v05/}{../experiments/repo_agent_v10/results/formal-main24-v2/postrun/publication-figures-v13-final/}}
\newcommand{\repair}{\ensuremath{\alpha}}
\newcommand{\regress}{\ensuremath{\beta}}
\newcommand{\risk}{\operatorname{Risk}}
\newcommand{\coverage}{\operatorname{Cov}}

\newif\ifcapableresults
\IfFileExists{capable-v14-results.tex}{%
  \capableresultstrue

}{%
  \capableresultsfalse
  
}
\IfFileExists{qwen3next-v1-results.tex}{%

}{%
}
\IfFileExists{devstral-v1-results.tex}{%

}{%
}
\IfFileExists{contract-components-v1-results.tex}{%

}{%
}
\IfFileExists{loopdepth-v1-results.tex}{%

}{%
}

\title{Looping Is Not Reliability: State-Bound Evidence and Typed Revision Contracts for Agentic Code Repair}

\author{Xueping Gao}
\correspondingauthor
\orcid{0009-0004-3096-2646}
\email{xueping.gxp@alibaba-inc.com}
\affiliation{
  \institution{Alibaba Cloud Computing}
  \city{Hangzhou}
  \country{China}
}

\author{Jianwei Yang}
\orcid{0009-0003-8035-0755}
\email{jianwei.yjw@alibaba-inc.com}
\affiliation{
  \institution{Alibaba Cloud Computing}
  \city{Hangzhou}
  \country{China}
}

\author{Qiang Yang}
\orcid{0009-0002-7522-2321}
\email{yuanzhan.yq@taobao.com}
\affiliation{
  \institution{Alibaba Cloud Computing}
  \city{Hangzhou}
  \country{China}
}

\renewcommand{\shortauthors}{Gao, Yang, Yang}

\begin{abstract}
Generate--test--revise loops are common in coding agents, but repetition alone provides no reliability guarantee. We study the gap between finding a correct patch and retaining, verifying, and submitting it. A sealed five-seed study over 30 HumanEval repairs produces 900 three-revision trajectories. Under forced revision, current correctness with current traces falls from 0.820 after one revision to 0.673 after two, although ever-correct rises to 0.847. Two common-state studies use 2,430 branches from identical frozen programs to remove post-treatment risk-set bias. In a prespecified 14B replication, stale traces harm 34/135 correct starts versus 4/135 with current traces, a 22.2-point increase (task-cluster 95\% CI $[8.9,37.0]$, exact Holm $p=0.0337$). A prospective 540-rollout policy eliminates observed correct-start harm but reduces wrong-start repair and fails its joint criterion. Repository experiments over 24 bugs and four coder stacks expose floor effects and component heterogeneity without Holm-significant effects. We therefore separate admission, preservation, grounded certification, competence, and liveness. We derive an evidence-bound typed loop contract and instantiate its mechanically enforceable subset in a reference implementation that binds verifier evidence to exact code states, preserves verified checkpoints, and emits auditable admission receipts. The implementation is an executable specification and conformance artifact, not evidence of improved repair competence or calibrated verifier dependence.
\end{abstract}



\begin{document}
\maketitle

\section{Introduction}

Coding agents increasingly operate as loops: propose code, execute tests, consume feedback, revise, and stop. Iterative refinement can improve the probability that a correct solution is produced \cite{madaan2023selfrefine,shinn2023reflexion,dai2025feedbackeval}. This search-centric view, however, obscures a deployment question: after a correct patch appears, what prevents the next revision from destroying it, a stale observation from reviving an old bug, or a correlated verifier from confirming the same error?

We call this distinction \emph{proposal search versus completion reliability}. Search asks whether any revision in a trajectory is correct. Completion reliability asks whether the orchestrator retains a correct state, bases revisions on evidence about that state, and stops only when its accepted evidence has sufficiently low risk. The distinction matters because an increasing \emph{ever-correct} rate can coexist with a decreasing current or final correctness rate.

\begin{figure}[t]
  \centering
  \includegraphics[width=\columnwidth]{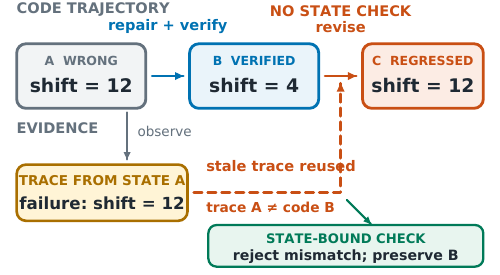}
  \caption{Mechanism of stale-evidence resurrection. Repair changes code state A (shift 12) to verified state B (shift 4), but an old trace bound to A is reused for the next revision and restores shift 12. State binding rejects the A/B mismatch before revision. Section~\ref{sec:stale-resurrection} reports HumanEval/89.}
  \Description{A two-lane flowchart. The solid code trajectory moves from wrong state A with shift 12 to verified state B with shift 4, then an unguarded revision regresses to state C with shift 12. A dashed evidence path carries a failure trace observed on state A into the revision of state B, explicitly marked trace A not equal to code B. A state-binding check rejects the mismatch instead.}
  \label{fig:stale-flow}
\end{figure}

Prior work establishes important pieces of this problem. Intrinsic self-correction may fail or regress without reliable external feedback \cite{huang2024cannot,kamoi2024when}; error localization and revision ability are distinct \cite{tyen2024errors}; and weak verifiers limit self-correction \cite{zhang2024verifiers}. FeedbackEval measures how feedback types affect code repair \cite{dai2025feedbackeval}. Work on evaluator self-preference and correlated model errors cautions that a second model is not automatically an independent judge \cite{panickssery2024evaluators,kim2025correlated}. These findings do not yet provide a joint account of state transitions, evidence provenance, verifier selective risk, verifier dependence, and stopping in one code-agent protocol.

This paper asks five research questions:
\begin{description}[leftmargin=0.8cm,style=nextline]
  \item[RQ1] Is correctness an absorbing state under repeated revision?
  \item[RQ2] From the same wrong or correct program, how do evidence content and state alignment change repair and continuation harm?
  \item[RQ3] Can verifier quality or independence be inferred from model-family diversity?
  \item[RQ4] Which orchestration guards directly block the observed failure classes, and what do they cost?
  \item[RQ5] How do contracts and revision allowance affect repository completion?
\end{description}

Our contributions are:
\begin{enumerate}[leftmargin=*,nosep]
  \item \textbf{A trajectory-level reliability decomposition} that jointly reports correctness discovery and retention, wrong-to-correct and correct-to-wrong transitions, verifier risk--coverage, conditional false-accept dependence, and sound completion.
  \item \textbf{Controlled evidence studies.} Forced revision shows correctness loss; paired common-state interventions at two coder capacities estimate state-alignment contrasts from identical selected states.
  \item \textbf{An executable contract and reference implementation.} We derive testable obligations for state-bound evidence, typed actions, checkpoint preservation, fresh completion certification, and risk-aware stopping. A reference implementation instantiates the mechanically enforceable subset as an admission layer around existing coding agents and verifiers.
  \item \textbf{Repository-agent factorials.} Over 24 bugs, four coder stacks test a contract; two Devstral factorials separate components and depth. All use sealed adjudication and task-cluster inference.
\end{enumerate}

\section{Related Work}

\subsection{Self-Correction and Feedback-Driven Repair}

Huang et al. find that intrinsic self-correction does not reliably improve reasoning when no external signal is available \cite{huang2024cannot}. Kamoi et al.'s survey similarly identifies reliable external feedback as a central success condition \cite{kamoi2024when}. Tyen et al. show that models often fail to locate errors even when they can correct an error whose location is supplied \cite{tyen2024errors}; Zhang et al. separate refinement ability from the strength of the verifier that decides when to refine \cite{zhang2024verifiers}. In code, FeedbackEval evaluates feedback types and reports diminishing gains over repeated repair \cite{dai2025feedbackeval}. We do not claim the new observation that feedback quality matters. We instead measure the bidirectional transition process and intervene on whether evidence is attached to the program state it describes.

Self-Refine and Reflexion exemplify inference-time refinement with generated feedback or memory \cite{madaan2023selfrefine,shinn2023reflexion}. Their motivating metric is typically task success. Our focus is the reliability contract required after a proposal exists: preservation, provenance, verification, and stopping.

\subsection{Agentic Software Engineering Systems}

Repository-scale systems make orchestration concrete. SWE-agent shows that the agent--computer interface changes coding performance \cite{yang2024sweagent}; Agentless reports that a simpler localization--repair--validation pipeline can outperform more elaborate open-source agents on SWE-bench Lite \cite{xia2025agentless}; and SWE-bench exposes long-context, multi-file demands \cite{jimenez2024swebench}. Beyond their end-task metrics, we test whether loops preserve valid states, consume state-aligned evidence, and gate completion: first under controlled function-level states, then as a bundle in repository Git/search/edit/test loops.

Recent process studies characterize feedback integration \cite{bouzenia2025trajectories}; the ``lucky pass'' AgentLens identifies regression cycles \cite{sahoo2026agentlens}, while a production-assessed AgentLens studies verification and recovery over whole trajectories \cite{podivilov2026agentlens}. Others control hint depth while measuring preservation \cite{le2026pairbench}, or expose action bias and visible--held-out gaps \cite{gloaguen2026fixedbench,zhao2026specbench}. Evidence guards, abstention, and transactional admission are established ideas \cite{meng2026eviact,cambronero2026abstain,chang2026mnemosyne}. Our narrower contribution is a fixed-code-state intervention on evidence provenance, linked to verifier risk and dependence and followed by a prospective policy test.

Adjacent infrastructure enforces related but different boundaries. In-toto records and verifies software-supply-chain steps \cite{torresarias2019intoto}, SLSA provenance describes where, when, and how an artifact was produced \cite{slsa2026provenance}, and OPA separates general policy decisions from their enforcement \cite{opa2026}. Our reference implementation adopts provenance and admission ideas but applies them inside an agent-repair transaction: verifier evidence is bound to the candidate code state on which it was produced, and admission preserves or recertifies that state. It neither provides supply-chain attestation nor repairs an inadequate oracle.

\subsection{Code Correctness and Executable Evidence}

HumanEval introduced execution-based functional correctness for generated programs \cite{chen2021humaneval}. EvalPlus expands its tests and demonstrates that a small visible suite can overestimate correctness \cite{liu2023evalplus}. We use disjoint visible, challenge, and hidden input sets derived from HumanEval+, while explicitly treating challenge and hidden tests as construction-dependent rather than independent oracles.

\subsection{Verifier Risk and Dependence}

LLM evaluators may recognize and favor their own generations \cite{panickssery2024evaluators}. More broadly, errors remain correlated across model architectures and providers \cite{kim2025correlated}. We therefore distinguish two quantities: selective risk at a specified coverage and conditional error dependence among incorrect candidates. Risk--coverage follows selective prediction, where abstention is part of evaluation rather than silently discarded \cite{geifman2019selectivenet}. Our contribution is to connect these quantities to agent completion decisions and to show that dependence changes with the acceptance operating point.

\section{Reliability Decomposition}

Consider the observed transition sketched in Figure~\ref{fig:stale-flow}. After the shift is repaired, the current program is correct, so $Y_t=1$. A failure trace produced from the earlier wrong program is then supplied as if it described the current state; the next revision restores the old defect, yielding $Y_{t+1}=0$. This is a correct-to-wrong transition. Conversely, changing a wrong program into a correct one is a wrong-to-correct transition. The following decomposition measures both rather than treating correctness as an absorbing endpoint.

Let $Y_t\in\{0,1\}$ denote the executable correctness of the code after revision $t$, and let $e_t$ denote the evidence supplied for the next revision. For a fixed evidence condition $e$, define
\begin{align}
\repair_e &= P(Y_{t+1}=1\mid Y_t=0,e),\\
\regress_e &= P(Y_{t+1}=0\mid Y_t=1,e).
\end{align}
If $q_t=P(Y_t=1)$, then
\begin{equation}
q_{t+1}-q_t=(1-q_t)\repair_e-q_t\regress_e.
\label{eq:transition}
\end{equation}
Repair ability alone is therefore insufficient: when the current state is likely correct, a nonzero regression rate can make another revision harmful in expectation. If $\repair_e+\regress_e>0$, continuation has negative expected correctness change when $q_t>\repair_e/(\repair_e+\regress_e)$.

This transition view makes the paper's three factors precise without pretending that all are separately identifiable. Let $e_0$ be a matched nonactionable control. We operationalize the \emph{effective evidence response} on wrong and correct states as
\begin{align}
R_W(e)&=\repair_e-\repair_{e_0}, &
R_C(e)&=\regress_{e_0}-\regress_e.
\label{eq:response}
\end{align}
These contrasts combine two latent quantities: how much task-relevant information the evidence carries and how responsively the coder maps that information into an edit. A single-coder evidence intervention identifies their joint effective action, not ``informativeness'' and ``responsiveness'' separately. Zero response can mean uninformative evidence or an unresponsive reviser; negative response can mean misleading or state-misaligned evidence, overreaction, or both. This identification limit is important: the three-factor thesis is an interface decomposition, not a fitted multiplicative law.

Evidence also has provenance. Write $e=(h_s,h_q,x)$ for source-state hash, test-suite hash, and payload. The evidence is state-aligned only if $h_s=h(S_t)$. Payload $x$ may contain true input/output facts while the envelope is invalid for the current state. The common-state study varies this alignment while holding the starting code fixed.

For verifier $v$ with threshold chosen to attain coverage $c$, we report
\begin{align}
\coverage_v(c)&=P(A_v=1),\\
\risk_v(c)&=P(Y=0\mid A_v=1),
\end{align}
where $A_v$ denotes acceptance. To avoid conflating shared prevalence with shared errors, dependence is computed only among incorrect candidates. For verifiers $i,j$ we report excess joint false acceptance,
\begin{equation}
D_{ij}(c)=P(A_iA_j\mid Y=0)-P(A_i\mid Y=0)P(A_j\mid Y=0),
\end{equation}
and the corresponding $\phi$ coefficient. Both are operating-point conditional: changing coverage changes which errors are accepted.

For an ensemble that accepts only when both verifiers accept, let $f_i=P(A_i=1\mid Y=0)$. Its conditional false-accept rate is
\begin{equation}
P(A_iA_j=1\mid Y=0)=f_if_j+D_{ij}.
\label{eq:joint-fa}
\end{equation}
Thus the familiar product-of-errors argument is valid only when $D_{ij}=0$. Positive dependence is an additive reliability penalty even if both marginal verifiers are individually strong; low dependence is not useful when either marginal risk is poor.

Finally, \emph{sound completion} is $P(A=1,Y=1)$ over all trajectories. Unlike risk alone, it exposes a policy that obtains zero observed error by rejecting nearly everything. We report coverage, risk, sound completion, and mean selected revision together rather than choosing a universal scalar utility.

\section{Study Design}

\subsection{Tasks and Executable Adjudication}

We select 30 Python function-repair tasks from the HumanEval portion of FeedbackEval \cite{dai2025feedbackeval}. Initial wrong candidates are stratified by source (existing benchmark candidate, LLM-generated candidate, and rule-based mutation) and frozen by hash before the confirmatory runs. We do not redistribute the source benchmark data; the artifact records task identifiers, candidate indices, sources, and hashes.

Correctness requires passing visible, challenge, and hidden tests. Challenge and hidden inputs are disjoint partitions of HumanEval+ \cite{liu2023evalplus}; they share the same benchmark construction process. Generation sees only visible evidence. Challenge and hidden suites remain sealed until all trajectory and branch matrices pass structural validation. Evaluation is cached by $(\text{suite},\text{task},\text{code SHA-256})$.

\subsection{Trajectory Study}

The coder is Qwen2.5-7B-Instruct \cite{qwen2024qwen25}, sampled with temperature 0.3 and top-$p$ 0.9. Each trajectory performs three forced revisions. We cross 30 tasks, six evidence conditions, and five seeds, yielding 900 trajectories and 2,700 revisions. The evidence conditions are current binary outcome, current execution trace, stale binary outcome, stale trace, unavailable evidence, and a flipped-binary stress condition. Forced revision intentionally estimates what happens when an orchestrator continues; it is not presented as the behavior of every adaptive agent.

Every revision is judged by native self-declaration, fresh-context Qwen2.5-7B, same-family Qwen2.5-14B, cross-family DeepSeek-Coder-6.7B \cite{deepseekcoder2024}, and an executable challenge verifier. This produces 13,500 verdicts. Replaying four stopping rules for each verifier produces 18,000 paired outcomes: verifier-only, visible gate, challenge gate, and a strict gate requiring challenge acceptance plus evidence sourced from the immediately preceding state.

The three model verifiers use an order-swapped, single-token A/B forced choice; the acceptance score is the mean ACCEPT probability under both label orders. Native thresholds are fixed at 0.5. Retrospective coverage matching ranks scores on the frozen revision distribution and accepts the top fraction; it compares verifiers but does not estimate deployment calibration. A stopping replay selects the first revision satisfying its gate; if none does, it rejects after revision 3. Because generation is frozen before replay, every policy sees the same candidates and executable outcomes. ``Mean rounds'' includes rejected trajectories, which consume the full three-revision budget.

\begin{table*}[t]
\caption{Evidence interventions. Current evidence is generated from the code being revised; stale evidence is generated from a different revision or the frozen wrong state.}
\label{tab:evidence}
\centering
\small
\begin{tabular}{llll}
\toprule
Condition & Payload & Source state & Role\\
\midrule
Current binary & Visible PASS/FAIL & Current code & Minimal aligned signal\\
Current trace & Outcome, pass count, up to 3 failures & Current code & Actionable aligned signal\\
Stale binary/trace & Same formats & Previous revision & Free-running freshness stress\\
Unavailable & \texttt{UNAVAILABLE} & None & No-evidence baseline\\
Flipped binary & Negated visible PASS/FAIL & Current code & Misleading-signal stress\\
Matched control & Neutral text, character-length matched & None & Report-length/attention control\\
Stale wrong trace & Failure trace & Frozen wrong code & State-misalignment intervention\\
\bottomrule
\end{tabular}
\end{table*}

\subsection{Common-State Intervention}

Transition rates in a free-running trajectory have a post-treatment risk-set problem: evidence affects whether a revision reaches a correct state, so comparing later regressions conditions on an evidence-dependent event. We address this with common-state intervention. For each task we freeze one wrong and one correct code state, selected before assigning the new evidence branches. Twenty-seven of 30 tasks contain both states and enter this study; the estimand therefore concerns tasks with an observed repairable state.

From every frozen wrong state we branch current binary, current trace, unavailable, and length-matched nonactionable evidence. From every correct state we branch the same conditions plus a stale trace generated from the paired wrong state. The prompt permits returning unchanged code. Crossing 27 tasks, nine state--evidence branches, and five common-random-number seeds gives 1,215 single-step branches for Qwen2.5-7B. Before inspecting replication outcomes, we freeze an exact Qwen2.5-14B repeat with the same states, evidence, prompts, sampling parameters, seeds, and adjudication, adding 1,215 branches. The two capacities are analyzed separately rather than pooled.

The frozen wrong state is the original visible-failing candidate. The frozen correct state is selected mechanically as the lexicographically first v0.4.3 revision passing all three suites; no failure case is hand-picked. The matched control contains no task, test, PASS/FAIL, or numerical semantics and exactly matches the target report in Unicode character count. It does not exactly match tokenizer length, so prompt-token distributions remain an audited nuisance variable. Within each task--state--seed block, evidence branches share the sampling seed; invalid or unchanged outputs remain in the intention-to-treat denominator.

\begin{table}[t]
\caption{Confirmatory study inventory.}
\label{tab:inventory}
\centering
\small
\begin{tabular}{lrr}
\toprule
Component & Trajectory study & Common-state study\\
\midrule
Tasks & 30 & 27\\
Seeds & 5 & 5\\
Coder capacities & 1 & 2\\
Evidence conditions & 6 & 4 wrong / 5 correct\\
Revisions per branch & 3 & 1\\
Trajectories/branches & 900 & 2,430\\
Candidate revisions & 2,700 & 2,430\\
Verifier verdicts & 13,500 & ---\\
Policy outcomes & 18,000 & ---\\
\bottomrule
\end{tabular}
\end{table}

\subsection{Inference and Audit}

The task, not the revision, is the independent sampling unit. Confirmatory common-state contrasts average seeds within each task and then weight tasks equally. Within each coder, we use 10,000 task-cluster bootstrap draws, 100,000 paired sign flips, and one Holm correction over six preregistered contrasts. Because five-seed task effects lie on a discrete grid, an exact dynamic-programming sign-flip sensitivity removes Monte Carlo error. A secondary paired analysis subtracts 7B task-level effects from 14B effects and applies a separate six-test Holm family; it was specified before inspecting 14B outcomes but is not a randomized model-family comparison. Trajectory confidence intervals also resample complete task clusters, preserving all seeds, conditions, revisions, and policies. Invalid generations remain in the intention-to-treat denominator. All protocol, data-partition, trajectory, verdict, and analysis artifacts carry SHA-256 lineage hashes.

Generation processes can read only the visible bundle. Challenge and hidden bundles are opened only after trajectory IDs, condition matrices, hashes, parser outputs, and lineage fingerprints validate exactly. Hidden adjudication then recomputes the complete verifier and policy matrices. This ordering prevents a partial run or adaptive policy choice from leaking hidden outcomes back into generation.

\subsection{Prospective Paired Online Commit-Admission Comparison}

Replay can show whether a guard would intercept recorded outputs, but not how enforcing it changes subsequent model behavior. We therefore run a sealed paired online comparison on the 27 common-state tasks: two start states, five seeds, and baseline/guarded arms yield 540 rollouts. Both arms use Qwen2.5-7B, the same typed action prompt, visible diagnostics, challenge executions, seed schedule, and three-step budget. Baseline commits every parseable patch and ignores challenge verdicts for admission. Guarded execution binds evidence to the current state, admits a patch only after the challenge gate, and retains a last-known-good checkpoint after invalid or rejected proposals. Hidden outcomes are opened only after the complete paired matrix validates. The preregistered endpoints are unsafe-completion superiority and sound-completion non-inferiority at a $-5$-point margin; inference uses paired task-cluster bootstrap intervals.

\subsection{Repository-Agent Factorial}

To test the contract beyond single-function repair, we freeze 24 historical bugs from nine Python projects, with at most three tasks per project. Eligibility requires direct-parent bug/fix endpoints, stable three-repeat signatures across the frozen suites, no ignored oracle, and a gold production patch of at most five files and 200 changed lines. The primary-only manifest excludes pilot tasks, is frozen before model output, and permits no outcome-driven replacement. Each task has a sanitized agent mirror containing only the buggy tree and a separate evaluator mirror containing sealed tests but no gold source object. A hash-locked local Qwen2.5-14B snapshot operates through typed search, view, edit, shell, test, submit, and escalate actions in a real Git worktree, sampled at temperature 0.3, top-$p$ 0.9, and at most 1,024 new tokens per action. We cross current versus one-step-stale diagnostic evidence with contract OFF versus ON and repeat every cell over three preregistered seeds, yielding $24\times2\times2\times3=288$ intention-to-treat rollouts. Each rollout is limited to 15 actions, three source revisions, six visible tests, and a shared wall-clock budget.

Contract ON bundles evidence--state binding, a private commit gate, rejection with rollback, last-known-good checkpoints, and completion certification. Contract OFF executes and records the same private gate but does not use its verdict to block a revision. Diagnostic, gate, fail-to-pass, pass-to-pass, and full-suite assets are content-hashed; generation can access only the diagnostic interface. Every proposed patch, including rejected proposals, is preserved and later replayed in an evaluator-owned worktree. Before unsealing, a 24-task by four-suite baseline audit found residue only from terminal full-suite runs; the frozen evaluator orders fail-to-pass, pass-to-pass, then full, and hard-resets and cleans before each candidate. Final resolution requires all three suites to pass. An \emph{orchestrator-accepted completion} is either an explicit agent submission or, in the ON arm, a checkpoint accepted by the frozen completion rule. Unsafe and sound completion are accepted-and-incorrect and accepted-and-correct, respectively.

The task is the inferential unit: seeds are averaged within task and all 24 tasks receive equal weight. The co-primary intersection--union decision requires the one-sided 95\% upper bound on the Contract ON-minus-OFF unsafe-completion effect to be below zero and the lower bound on sound completion to exceed the preregistered absolute $-5$-point margin. This margin was frozen before outcomes; when baseline sound completion later falls below 5\%, its non-inferiority decision is necessarily weak evidence about liveness. We use 200,000 task-cluster bootstrap draws; task sign flips, project-equal estimates, and leave-one-project-out checks are sensitivities.

To probe stack generality under the 14B completion floor, we froze three separate
288-rollout replications before their outcomes: Qwen2.5-Coder-32B-Instruct-AWQ,
Qwen3-Coder-Next-FP8, and Devstral Small 2 24B. Each preserves the task matrix,
co-primary estimands, budgets, completion rule, and task-equal inference, and is
analyzed separately. Model, tokenizer, adapter, and runtime changes remain
bundled ecological contrasts, not identified capacity or family effects.
Outcome-blind runtime or adapter faults were repaired before restarting empty
main roots; discarded attempts remain receipted and no scientific design changed.

With Devstral fixed, a 576-rollout $2^3$ factorial randomizes state binding,
proposal admission/rollback, and completion certification. Task-equal main
effects enter one six-test Holm family over unsafe and sound completion. A
separate 576-rollout equal-budget factorial crosses Contract OFF/ON with revision
limits 1, 3, 5, and 8 while holding action, test, token, and wall budgets fixed;
its six-test family covers depth, contract, and interaction effects. Full stack,
runtime, pilot, and sensitivity specifications are in the reproduction artifact.

All reported experiments use the frozen research harness described above. The reference implementation was developed afterward as an executable realization of the mechanically enforceable contract obligations and was not used to generate the reported empirical results.

\section{Results}

\subsection{RQ1: Correctness Is Non-Absorbing}

\begin{figure*}[t]
  \centering
  \includegraphics[width=0.96\textwidth]{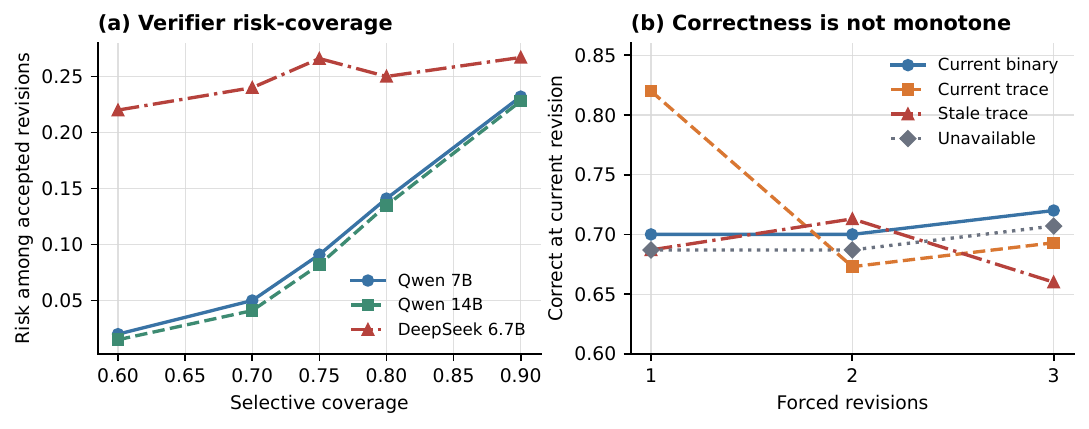}
  \caption{Left: retrospective selective risk at matched coverage. Cross-family diversity does not ensure verifier quality. Right: current correctness across forced revisions is non-monotone even while cumulative ever-correct increases.}
  \Description{Two line charts. The left plots risk versus coverage for Qwen 7B, Qwen 14B, and DeepSeek 6.7B. The right plots current correctness over three revisions for four evidence conditions.}
  \label{fig:depth-risk}
\end{figure*}

Figure~\ref{fig:depth-risk}b separates current correctness from cumulative discovery. With current traces, 82.0\% of trajectories are correct after revision 1. Current correctness then falls to 67.3\% at revision 2 and recovers only to 69.3\% at revision 3. In contrast, ever-correct increases from 82.0\% to 84.7\% and 85.3\%. At revision 3, 16.0\% of trajectories have produced a correct patch and subsequently lost it. Current binary evidence has lower loss (8.0\%) but still does not make correctness absorbing.

The shape is evidence-specific rather than a generic depth penalty. Current-binary correctness is $0.700,0.700,0.720$ across revisions; stale-trace correctness is $0.687,0.713,0.660$; and unavailable evidence is $0.687,0.687,0.707$. Consequently, neither ``more rounds help'' nor ``more rounds hurt'' is supported as a context-free rule. Loop depth changes the number of transition opportunities; the transition kernel determines their sign.

This result is deliberately scoped: it establishes non-absorption under the tested forced-revision protocol. It does not show that an adaptive loop must regress. It shows why an adaptive loop needs a preservation and stopping rule.

\subsection{RQ2: Evidence Helps Repair but Can Harm Retention}

\begin{figure}[t]
  \centering
  \includegraphics[width=\columnwidth]{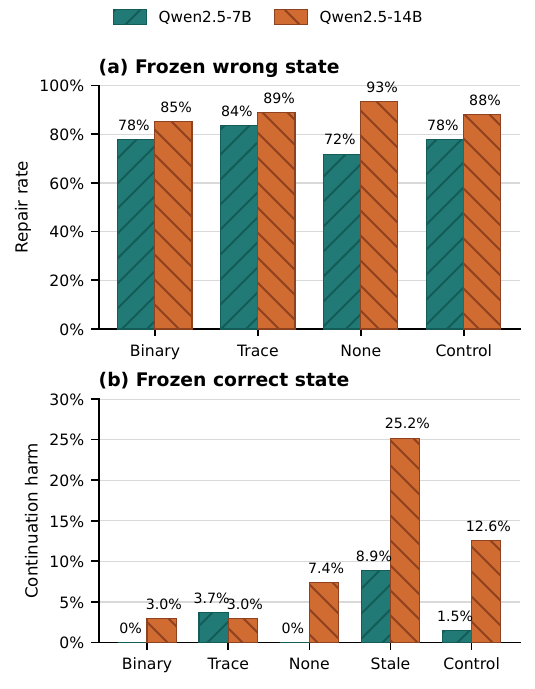}
  \caption{Assignment outcomes from identical frozen states. Evidence ordering changes across coder capacities; 14B stale evidence produces the largest continuation harm. Inference is reported in text.}
  \Description{Two grouped bar charts compare Qwen 7B and 14B repair probabilities from wrong states and continuation harm probabilities from correct states.}
  \label{fig:common}
\end{figure}

Table~\ref{tab:common} and Figure~\ref{fig:common} consolidate the assignment outcomes. For 7B, current traces produce the most repairs from wrong starts (113/135), while stale traces produce the most harms from correct starts (12/135). Trace minus control repair is $+5.9$ points with task-cluster 95\% CI $[-2.2,17.0]$; all six 7B contrasts have Holm-adjusted $p=1.0$.

The 14B replication changes the ordering: unavailable evidence produces the most repairs (126/135), so additional task feedback is not monotonically useful. From correct starts, stale traces harm 34/135 branches versus 4/135 under current traces, a $+22.2$-point contrast with CI $[8.9,37.0]$ and exact Holm $p=0.0337$. Stale minus control is $+12.6$ points, CI $[0.7,25.9]$, but does not survive Holm correction ($p=0.3934$). Thus long-context editing pressure explains part, but not all, of the stale-current gap.

The stale-minus-current interaction across coders is $+17.0$ points (task-cluster 95\% CI $[0.7,33.4]$) but remains suggestive rather than confirmatory after its separate Holm correction ($p=0.3615$); we therefore do not claim a general capacity effect. For 14B correct starts, stale traces change code in 81.5\% of branches, versus 47.4\% for current traces and 77.8\% for control; the corresponding harms are 25.2\%, 3.0\%, and 12.6\%. Churn is therefore neither repair nor semantic harm. Under both coders, character matching leaves a token-length difference (498.4 versus 451.7 mean prompt tokens for stale and control), an audited nuisance rather than a complete explanation.

\begin{table}[t]
\caption{Common-state repair and continuation harm. Each cell is event count/135; starts and assignments are identical across coders.}
\label{tab:common}
\centering
\small
\begin{tabular}{llrr}
\toprule
Start & Evidence & 7B event & 14B event\\
\midrule
Wrong & Current binary & 105/135 & 115/135\\
Wrong & Current trace & 113/135 & 120/135\\
Wrong & Unavailable & 97/135 & 126/135\\
Wrong & Matched control & 105/135 & 119/135\\
\midrule
Correct & Current binary & 0/135 & 4/135\\
Correct & Current trace & 5/135 & 4/135\\
Correct & Unavailable & 0/135 & 10/135\\
Correct & Stale wrong trace & 12/135 & 34/135\\
Correct & Matched control & 2/135 & 17/135\\
\bottomrule
\end{tabular}
\end{table}

\subsection{Stale Evidence Resurrects Fixed Bugs}
\label{sec:stale-resurrection}

\begin{figure}[t]
  \centering
  \includegraphics[width=\columnwidth]{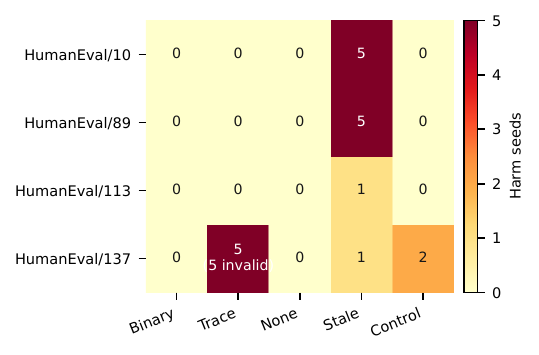}
  \caption{7B correct-to-incorrect outcomes are clustered, not uniform. HumanEval/10 and /89 show stale-evidence bug resurrection in 5/5 seeds; current-trace failures on /137 are invalid action-schema outputs.}
  \Description{Heatmap of four affected tasks by five evidence conditions, with counts of harmful seeds and invalid outputs.}
  \label{fig:clusters}
\end{figure}

Average uncertainty does not erase constructive counterexamples. For 7B, all correct-state harms occur in 4/27 tasks (Figure~\ref{fig:clusters}); stale traces regress HumanEval/10 and /89 in all five seeds. In /89, a stale report from code using Caesar shift 12 changes the fixed shift 4 back to 12 in every seed. The report contains valid input/expected-output facts, but falsely asserts that they describe the current code.

For 14B, stale harm is broader: 34 branches across 12/27 task clusters. HumanEval/10 and /46 regress in 5/5 stale seeds while their matched controls harm 0/5 and 1/5, respectively. Conversely, /89 no longer regresses, and /50 and /120 fail under both stale and control prompts. This redistribution is important: stale evidence is a reproducible failure mechanism, not a deterministic property of a task, and responsiveness determines whether the mechanism fires.

This yields a sharper mechanism than ``more information helps'': utility depends jointly on payload, provenance, and the coder's response. Candidate-source diagnostics also shift across coders and were not randomized, so they are descriptive rather than subgroup effects. The clustering and redistribution explain why task-level, not branch-level, uncertainty is essential.

\subsection{RQ3: Verifier Quality Is Not Independence}

At native thresholds, Qwen-7B covers 0.757 of revisions at risk 0.095; Qwen-14B covers 0.711 at risk 0.040; DeepSeek covers 0.991 at risk 0.298. Coverage confounds direct comparison, so Figure~\ref{fig:depth-risk}a imposes retrospective matched coverage. At 70\%, risks are 0.050, 0.041, and 0.240, respectively. Qwen-14B minus DeepSeek is $-19.9$ points with task-bootstrap CI $[-26.4,-9.5]$ when each bootstrap draw refits the cutoff. This compares the frozen distribution; deployment requires held-out calibration.

This ordering is not an artifact of choosing 70\% coverage: Qwen-14B minus DeepSeek risk differences remain negative from 60\% through 90\% coverage, ranging from $-20.5$ to $-4.0$ points, with all five task-bootstrap intervals below zero. Repeated code states do affect the magnitude. Collapsing 2,700 revisions to 617 unique task--code hashes raises 70\%-coverage risks to 0.400, 0.387, and 0.472; Qwen-14B still has lower empirical AURC (0.225 versus 0.402) and 8.5 points lower risk than DeepSeek. We therefore interpret 19.9 points as the revision-distribution estimate, not an intrinsic model constant.

Quality also does not imply independent errors. Among hidden-incorrect revisions at 70\% coverage, Qwen-7B/Qwen-14B have $\phi=0.641$ with task-cluster CI $[0.096,0.935]$ and excess joint false acceptance $0.060$ $[0.001,0.133]$. Qwen-14B/DeepSeek have $\phi=0.021$ $[-0.231,0.272]$ at the same coverage, but DeepSeek's risk is much higher. At 75\% coverage, all three scored pairs have positive excess-joint intervals. Dependence is thus an operating-point property, not a static label attached to a pair of model families.

\begin{table}[t]
\caption{Verifier quality and conditional dependence. Risk is at 70\% coverage; $\phi$ is conditional on incorrect revisions.}
\label{tab:verifiers}
\centering
\small
\begin{tabular}{lrr}
\toprule
Verifier/pair & Risk & Conditional $\phi$\\
\midrule
Qwen-7B & 0.050 & ---\\
Qwen-14B & 0.041 & ---\\
DeepSeek-6.7B & 0.240 & ---\\
\midrule
Qwen-7B / Qwen-14B & --- & 0.641\\
Qwen-7B / DeepSeek & --- & $-0.072$\\
Qwen-14B / DeepSeek & --- & 0.021\\
\bottomrule
\end{tabular}
\end{table}

\subsection{RQ4: Bundled Admission Changes Retention and Repair}

Table~\ref{tab:policy} evaluates stopping and verification on the same 900 trajectories. Self-only stopping accepts 99.8\% but has 29.5\% risk and 70.3\% sound completion. A visible gate prevents 248 false completions and introduces none, reducing risk to 2.4\% while increasing sound completion to 77.8\%. The challenge gate prevents all 265 self-only false completions in this sample and attains 77.8\% sound completion at 77.8\% coverage. Its zero observed errors do not imply zero population risk: the conservative one-sided 95\% zero-event bound over 30 task clusters is 9.5\%.

Task-cluster uncertainty is wide enough to matter. Self-only risk has a 95\% interval of $[0.154,0.445]$, visible-gated risk $[0,0.077]$, and visible-gated sound completion $[0.642,0.900]$. The zero-event task bound is $1-0.05^{1/30}=0.095$; treating 1,877 challenge-passing revisions as independent would instead give 0.0016 and would be anti-conservative for cross-task generalization. The empirical result supports executable gating in this domain, not a universal safety guarantee.

Strict freshness is not free. Qwen-14B plus a current-evidence challenge has no observed false completion but covers only 38.4\%, loses 311 baseline sound completions, and requires 2.27 mean revisions. Reliability evaluation must expose this abstention/cost trade-off.

The prospective paired comparison estimates the bundled policy's orchestrator-level effect. From correct starts, baseline continuation harms 13/135 branches and finishes correct on 122/135; guarded execution has no observed harm and retains correctness on 135/135. The $-9.6$-point harm contrast has a task-bootstrap 95\% CI of $[-20.7,0.0]$ and sign-flip $p=0.25$, because all baseline harms occur in only 3/27 task clusters. From wrong starts, repair falls from 100/135 to 93/135. Across both start states, unsafe completion changes from 1/270 to 0/270 (effect $-0.4$ points, CI $[-1.1,0.0]$), which is too sparse to demonstrate superiority. Sound completion changes from 219/270 to 211/270 (effect $-3.0$ points, CI $[-8.5,1.1]$), so the preregistered $-5$-point non-inferiority criterion is not met. The bundled guard therefore shifts observed errors while reducing both revision plasticity and completion coverage; these data do not identify any component's individual effect.

\begin{table}[t]
\caption{Stopping-policy replay over 900 fixed trajectories. CIs are task-clustered; paired changes are reported in text.}
\label{tab:policy}
\centering
\scriptsize
\begin{tabular}{lrrrr}
\toprule
Policy & Cov. & Risk & Sound & Rounds\\
\midrule
Self only & 0.998 & 0.295 & 0.703 & 1.03\\
Self + visible & 0.797 & 0.024 & 0.778 & 1.55\\
Self + challenge & 0.778 & 0.000 & 0.778 & 1.56\\
Qwen-14B only & 0.794 & 0.046 & 0.758 & 1.52\\
Qwen-14B + challenge & 0.769 & 0.000 & 0.769 & 1.55\\
Challenge executable & 0.788 & 0.000 & 0.788 & 1.52\\
Qwen-14B + fresh challenge & 0.384 & 0.000 & 0.384 & 2.27\\
\bottomrule
\end{tabular}
\end{table}

\subsection{RQ5: Repository Contracts Expose Completion-Gate Dependence}

\begin{figure*}[t]
  \centering
  \includegraphics[width=0.88\textwidth]{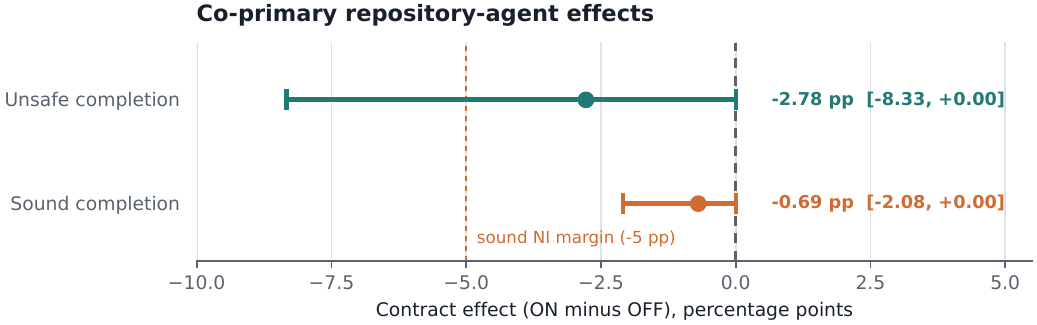}
  \caption{Prespecified repository-agent contract effects (ON minus OFF). Bars are task-cluster bootstrap 95\% intervals over 24 task-level effects; the three seeds per cell are averaged before inference. Unsafe-completion superiority requires its one-sided upper bound below zero; sound-completion non-inferiority requires its lower bound above $-5$ points. Only the latter, floor-limited criterion is met, so the joint IUT fails.}
  \Description{Forest plot with two contract effects. Unsafe completion decreases by 2.78 percentage points with a confidence interval from minus 8.33 to zero. Sound completion decreases by 0.69 points with an interval from minus 2.08 to zero, above the minus 5 point non-inferiority margin.}
  \label{fig:repo-primary}
\end{figure*}

Table~\ref{tab:repo-stacks} consolidates the four stack matrices. The original
Qwen2.5-14B study has a severe completion floor: Contract OFF produces five
accepted completions, whereas Contract ON produces none. It removes all four
observed unsafe completions but also the only sound one; 276/288 rollouts
exhaust the action budget. Figure~\ref{fig:repo-primary} gives the task-equal
effects. Unsafe completion changes by $-2.78$ points (task-cluster 95\% CI
$[-8.33,0.00]$), missing strict superiority, while sound completion changes by
$-0.69$ points ($[-2.08,0.00]$), passing the $-5$-point margin. With only one
OFF-arm sound completion, that pass is weak liveness evidence and the joint
decision is negative.

\begin{table*}[t]
\caption{Repository-agent stack matrices. Each arm has 72 rollouts and reports
accepted/unsafe/sound counts; effects are task-equal Contract ON minus OFF
percentage points with task-cluster 95\% CIs. Every joint IUT fails; sound
non-inferiority passes only under severe completion floors.}
\label{tab:repo-stacks}
\centering
\small
\setlength{\tabcolsep}{3pt}
\begin{tabular}{lccccrr}
\toprule
Stack & Current/OFF & Current/ON & Stale/OFF & Stale/ON & $\Delta$ unsafe [95\% CI] & $\Delta$ sound [95\% CI]\\
\midrule
Qwen2.5-14B & 3/2/1 & 0/0/0 & 2/2/0 & 0/0/0 & $-2.78\;[-8.33,0.00]$ & $-0.69\;[-2.08,0.00]$\\
Qwen2.5-Coder-32B & 2/0/2 & 5/3/2 & 1/0/1 & 4/3/1 & $+4.17\;[0.00,11.11]$ & $0.00\;[-2.08,2.08]$\\
Qwen3-Coder-Next & 0/0/0 & 2/0/2 & 0/0/0 & 4/0/4 & $0.00\;[0.00,0.00]$ & $+4.17\;[0.00,11.11]$\\
Devstral Small 2 24B & 0/0/0 & 2/0/2 & 0/0/0 & 2/0/2 & $0.00\;[0.00,0.00]$ & $+2.78\;[0.00,7.64]$\\
\bottomrule
\end{tabular}
\end{table*}

The 32B stack is the only replication with observed unsafe completion under
Contract ON: six unsafe completions are concentrated in two tasks, and the gate
false-passes incorrect proposals and submissions. Qwen3-Coder-Next and Devstral
show no observed unsafe completion, but they produce only six and four accepted
completions, respectively; 282/288 and 284/288 rollouts exhaust their action
budgets. These are stack-level ecological contrasts, not identified capacity or
model-family effects. Full gate, cost, and sensitivity breakdowns remain in the
artifact.

Table~\ref{tab:repo-factorials} moves the two Devstral factorials into one
inferential summary. No component, depth, contract, or interaction effect
survives its preregistered six-test Holm family. Certification has the largest
descriptive sound-completion effect, but only four tasks support it and $C=0$
removes the checkpoint-completion path; the contrast therefore partly reflects
endpoint mechanics rather than an isolated semantic verification benefit. The
depth factorial contains 0/576 unsafe and 25/576 sound completions. Revision
caps rarely bind, so its null depth result concerns this sparse-proposal process,
not loop depth in general.

\begin{table*}[t]
\caption{Devstral factorial effects (percentage points; $n=576$ each). CIs are
task-cluster 95\%; $p_H$ is Holm-adjusted within each six-test family. B, A,
and C denote state binding, admission/rollback, and completion certification.}
\label{tab:repo-factorials}
\centering
\small
\setlength{\tabcolsep}{4pt}
\begin{tabular}{llcc}
\toprule
Design ($n=576$) & Contrast & $\Delta$ unsafe [95\% CI]; $p_H$ & $\Delta$ sound [95\% CI]; $p_H$\\
\midrule
$2^3$ components & B: state binding & $-0.69\;[-2.08,0.00];\;1.000$ & $+0.69\;[0.00,1.74];\;1.000$\\
 & A: admission/rollback & $0.00\;[0.00,0.00];\;1.000$ & $+0.69\;[-0.69,2.43];\;1.000$\\
 & C: certification & $+0.69\;[0.00,2.08];\;1.000$ & $+6.25\;[0.69,13.19];\;0.753$\\
\midrule
$4\times2$ depth--contract & Cap 1 to cap 8 & $0.00\;[0.00,0.00];\;1.000$ & $-0.23\;[-1.77,1.09];\;1.000$\\
 & Contract ON minus OFF & $0.00\;[0.00,0.00];\;1.000$ & $+3.13\;[0.00,9.03];\;1.000$\\
 & Depth $\times$ contract & $0.00\;[0.00,0.00];\;1.000$ & $-0.45\;[-7.63,6.27];\;1.000$\\
\bottomrule
\end{tabular}
\end{table*}

Across the four stack matrices, accepted-completion totals are only 5, 12, 6,
and 4, and OFF-arm sound-completion rates are 0--2.1\%. Thus the tables expose
the common floor directly: the contract changes completion control but does not
establish competence or liveness.

\section{Evidence-Bound Typed Loop Contract}

The observed failures motivate an auditable orchestration specification, not a universally validated remedy or a new prompting recipe. Its four interfaces separate obligations that the negative policy and factorial decisions show must be evaluated rather than assumed.

\paragraph{State-bound evidence.}
Every observation is carried in an envelope
\begin{equation*}
E=(h_{code},h_{suite},execution\_id,payload).
\end{equation*}
The orchestrator may use $E$ to revise only when $h_{code}$ equals the current state hash. A mismatch triggers refresh, \textsc{Keep}, or \textsc{Escalate}; it never directly triggers a patch.

\paragraph{Typed revision actions.}
Free-form text conflates ``no edit is needed'' with ``the attempted patch was unparsable.'' We use
\begin{equation*}
RevisionAction=\textsc{Keep}\mid\textsc{Patch}(code)\mid\textsc{Escalate}(reason).
\end{equation*}
Only a parsed, executable \textsc{Patch} may replace the current state. In the 7B common-state study, all five current-trace harms on HumanEval/137 are responses containing only \texttt{STATUS: DONE} without code. A fail-closed parser that maps this action to \textsc{Keep}, rather than treating it as an invalid patch, would retain the correct checkpoint in these recorded branches.

\paragraph{Last-known-good checkpoint.}
The ledger records code and evidence hashes for the last state that passed the configured gate. Invalid actions never overwrite it. This makes correctness absorbing with respect to parser and protocol failures even when model behavior is not.

\paragraph{Risk- and dependence-aware stopping.}
Verifier selection uses risk at the intended coverage and conditional dependence, not model diversity as a proxy. Our matched-coverage estimates are retrospective; deployment additionally requires held-out calibration and drift monitoring. A completion decision records its operating point so residual risk remains auditable.

\begin{table*}[t]
\caption{Contract obligations, reference mechanisms, conformance cases, and residual risks.}
\label{tab:obligations}
\centering
\footnotesize
\begin{tabular}{p{0.19\textwidth}p{0.25\textwidth}p{0.20\textwidth}p{0.27\textwidth}}
\toprule
Obligation & Reference mechanism & Conformance evidence & Residual risk\\
\midrule
State-bound evidence & Bind candidate, tree, policy, verifier, and context identities & Reject replay, mutation, and wrong-tree evidence (SB001--SB007) & Aligned evidence may be wrong; host and verifier remain trusted\\
Independent admission & Broker alone advances verified state and issues decisions & Self-report cannot authorize admission (SB021, SB027) & Process separation does not imply error independence\\
Checkpoint preservation & Retain and exactly restore the last verified candidate & Regressions and malformed actions cannot overwrite it (SB008--SB010, SB014, SB016) & Only oracle-covered properties are preserved\\
Fresh certification & Re-execute checks on the selected completion state & Failed recertification cannot complete (SB012, SB020, SB031) & Flakiness, drift, and suite incompleteness remain\\
Auditable stopping & Typed outcomes, hash-linked events, digest-bearing receipt, coverage and liveness fields & False pass/rejection, abstention, and coverage collapse are explicit (SB028--SB035) & Auditability is not calibration, liveness, or attestation\\
\bottomrule
\end{tabular}
\end{table*}

\subsection{Operational Semantics}

The contract turns a permissive text loop into a small state machine. At step $t$, the orchestrator stores current state $S_t$, last-known-good state $L_t$ (possibly empty), and an append-only evidence ledger. It executes the following transaction:
\begin{enumerate}[leftmargin=*,nosep]
  \item run a named suite on $S_t$; append evidence bound to the state and suite;
  \item reject or refresh $E_t$ if either hash does not match the requested state and suite;
  \item request one typed action; \textsc{Keep} leaves state unchanged, \textsc{Escalate} abstains, and only a parseable \textsc{Patch} creates candidate $S'_t$;
  \item evaluate $S'_t$ with the configured commit gate; on failure retain $S_t$ or roll back to $L_t$, never overwrite $L_t$;
  \item stop only if the completion policy accepts the committed state at its declared risk--coverage operating point; otherwise continue within budget.
\end{enumerate}
Two safety properties follow mechanically. First, a state-hash mismatch cannot trigger a commit. Second, parser or action-schema failure cannot erase a checkpoint. Neither property claims that a hash-matched patch is semantically correct; that residual risk belongs to the executable and learned verifiers.

\subsection{StateSeal Reference Implementation and Conformance}

StateSeal wraps existing agents and project-defined verifiers. Its admission broker recomputes state identity, verifies an isolated proposal outside the agent process, preserves a checkpoint, and recertifies it before issuing a digest-bearing receipt. It records coverage and delivery impact but does not learn risk or dependence.

At the frozen snapshot, all 35 deterministic failure-injection cases and three managed-language fixtures pass. This is conformance evidence, not effectiveness, competence, verifier independence, or study validation. The public \href{https://github.com/hellogxp/stateseal/releases/tag/v0.1.1}{v0.1.1 release} provides source, commands, and the case catalog.

In deterministic replay, state binding intercepts all 12 and 34 stale-condition harms for 7B and 14B; adding typed action reduces total harm from $19\to2$ and $69\to33$. A challenge checkpoint blocks all 19 7B harms but misses two 14B hidden-test regressions. Because stale evidence is assigned only to correct starts, replay diagnoses mechanisms rather than end-to-end utility.

\section{Discussion}

\paragraph{Controlled counterexamples.}
Search does not ensure retention; valid reports become harmful under state mismatch; and model-family diversity ensures neither low verifier risk nor independent errors. Iteration cannot substitute for evidence, state management, or verification.

\paragraph{What they do not refute.}
Loops remain useful when expected repair exceeds continuation harm. We critique unconditional continuation and unqualified completion, not adaptive search: depth samples transitions; orchestration determines whether value accumulates.

\paragraph{Why the three factors are not interchangeable.}
Better feedback cannot help an unresponsive coder; responsiveness can amplify stale evidence; and proposal quality cannot offset correlated false acceptance. Equation~\ref{eq:joint-fa} requires both low marginal false acceptance and low dependence.

\paragraph{Safety, plasticity, liveness, and interface fit are separate.}
The prospective policy prevents observed regression but rejects useful repairs; repository contracts similarly trade unsupported against sparse sound completion. Neither randomized factorial is Holm-significant. Admission, preservation, certification, interface fit, and depth remain distinct.

\paragraph{Implications for agent evaluation.}
Pass@1 or best-of-$k$ misses state loss, incorrect acceptance, and low risk from abstention. Evaluations should also report current versus ever-correct, common-state transitions, coverage--risk, sound completion, cost, and conditional verifier overlap. Our 24 repository bugs test mechanisms, not prevalence.

\paragraph{Mechanism, policy, and calibration.}
State binding, transition guards, checkpoints, and recertification are mechanical. Verifier adequacy, error dependence, residual risk, and safety--liveness remain empirical. The implementation makes failure modes falsifiable; it does not improve agent or oracle competence.

\section{Threats to Validity}

\paragraph{External validity.}
The trajectory and common-state studies use public HumanEval tasks that may occur in model training data, and 27 tasks are selected for 7B-observed repairability. Pairing strengthens assignment contrasts but not absolute-rate generalization. The repository study covers 24 bugs from nine Python projects, four coder stacks, and one scaffold. Stack factors vary jointly, interface friction is material, and tasks within projects may be correlated; these are mechanism tests, not random samples of agents, languages, projects, or bugs.
The controlled function-level studies stop at 14B parameters and do not establish that frontier-scale hosted models have the same transition kernel.

\paragraph{Construct validity.}
Challenge and hidden inputs are disjoint but share HumanEval+ construction, and no finite repository suite is a complete specification. One-step staleness is an intervention, not its natural prevalence; exact-code change is not semantic churn. Verifier family and capacity are confounded. Self acceptance is declarative, whereas repository acceptance also includes the ON arm's frozen checkpoint rule.

\paragraph{Internal and statistical validity.}
Forced revision is not a natural adaptive policy. Common states address post-treatment risk-set bias, but all six 7B and five of six 14B contrasts are nonsignificant after Holm correction; no secondary cross-coder interaction survives correction. The original repository contract identifies only a joint effect; component randomization remains event-sparse, and the depth caps rarely bind. The four stack matrices have only 5, 12, 6, and 4 accepted completions; under this floor, the absolute $-5$-point non-inferiority margin weakly constrains liveness. Task-equal inference has project-equal and leave-one-project-out checks, but only nine projects support project-level generalization.

\paragraph{Reproducibility.}
Prompts, private suites, full lineage, and the numerical reconstruction package are retained internally. The public materials provide the StateSeal v0.1.1 source, commands, and case catalog; licensing and evaluator privacy prevent redistribution of the full raw trajectories and private suites. The post-study artifact tests mechanism conformance, not effectiveness, verifier adequacy or independence, or trust in the host, runner, Git, and OS. Strict admission can reject useful repairs; its UI is outside the trusted boundary.

\section{Conclusion}

Iteration supplies proposals, not reliability. The typed contract separates state alignment, admission, preservation, grounded completion, competence, liveness, and interface fit; its implementation enforces state binding, checkpoints, recertification, and auditable decisions. Verifier quality, dependence, and general reliability remain empirical.

\noindent\textbf{Availability and disclosures.} The numerical reconstruction package is retained internally; private suites and licensed source are not redistributed. StateSeal v0.1.1 is at \url{https://github.com/hellogxp/stateseal}. AI-assisted code and prose were author-verified.

\balance
\bibliographystyle{ACM-Reference-Format}
\bibliography{references}

\end{document}